# Deep Learning for Apple Diseases: Classification and Identification

Asif Iqbal Khan[a,*], SMK Quadri[b] , Saba Banday[c]

[a,b] Department of Computer Science, Jamia Millia Islamia, New Delhi, India
[c] Department of Pathology, Sher-i-Kashmir University of Agricultural Sciences, Shalimar, Srinagar, J&K

*Abstract*: - Diseases and pests cause huge economic loss to the apple industry every year. The identification of various apple diseases is challenging for the farmers as the symptoms produced by different diseases may be very similar, and may be present simultaneously. This paper is an attempt to provide the timely and accurate detection and identification of apple diseases. In this study, we propose a deep learning based approach for identification and classification of apple diseases. The first part of the study is dataset creation which includes data collection and data labelling. Next, we train a Convolutional Neural Network (CNN) model on the prepared dataset for automatic classification of apple diseases. CNNs are end-to-end learning algorithms which perform automatic feature extraction and learn complex features directly from raw images, making them suitable for wide variety of tasks like image classification, object detection, segmentation etc. We applied transfer learning to initialize the parameters of the proposed deep model. Data augmentation techniques like rotation, translation, reflection and scaling were also applied to prevent overfitting. The proposed CNN model obtained encouraging results, reaching around 97.18% of accuracy on our prepared dataset. The results validate that the proposed method is effective in classifying various types of apple diseases and can be used as a practical tool by farmers.

Keywords: Deep Learning, Apple disease classification, Convolutional Neural Network

## 1. Introduction

Apple, rice and saffron are three key crops of Kashmir valley. Among these three, Apple occupies a prominent place in its agricultural economy because of right agro-climatic conditions with comparative advantage "ecological niche" for apple farming. Apple dominates the agricultural economy of the valley with an annual productivity of around 180000 MTs (Horticulture Department, 2019), in which an enormous proportion is exported to different parts of the world. In Jammu and Kashmir (J&K), around 70% of total population depends directly or indirectly on agriculture and as regards the production quantity, about 75% of apple production in India comes from Kashmir. Currently, around 160000 hectares of land in the Valley is under apple cultivation and the number is going up as more and more people are converting their rice fields into Apple orchards as the fruit generates more revenue and is less labor intensive. The current production of apples stands at 12 MTs/hectare in Kashmir valley which is far below the level achieved by other countries (40 to 60 MT/ha). As per the experts, the main reasons for limited production are insects, crop diseases, pests, lack of proper disease detection and forecast system [1]. Diseases and pests cause huge economic loss to the apple industry every year. In July 2013, a disease called Alternaria unfurled in apple orchards of the valley. The disease spread like wildfire infecting more than 70% of the cultivars in the Baramulla and Bandipora districts. The disease resulted in extensive fruit fall and hence decrease in fruit production [2]. The same disease attack was again reported in 2018 [3]. According to the domain experts, one of the main reasons for the spread of this catastrophic disease was the lack of proper disease forecasting and detection system [2]. An on-time detection system would have detected the disease in the early stages and timely action could have avoided the damage. Alternaria is not the only threat to apple production. Like other crops, apples are also affected by a number of other diseases. Some of the common diseases identified by the experts and researchers are Scab, Canker, Apple Mosaic, Marssonina leaf blotch *(MLB),* Leaf spots, Brown Rot, wooly aphid, powdery mildew etc.

Pests and diseases not only reduce the fruit production but adversely affect the fruit quality as well. Thus, in order to improve the quality and quantity of crops, it is imperative to protect them from diseases. Identifying a disease correctly when it first appears and taking timely action is a crucial step for efficient disease management. Timely and accurate diagnosis of a disease attack is crucial for early and targeted application of curative measures and stopping the disease from spreading. Appropriate knowledge of a disease would help the farmer to take proper precautionary measures or apply just the right amounts of pesticides, thereby getting both economic and environmental benefits.

However, to achieve this, farmers must continuously monitor their crops and remain in touch with the experts for any help from time to time. The experts must be proficient and should have extensive knowledge of various diseases, their symptoms and treatment. Such methods are labor intensive and time-consuming. Moreover, some of the diseases can only be diagnosed by experts and only few samples can be examined at any time. A valid alternative to such a labor intensive and costly task will be an automatic system that can detect diseases in early stage and provide appropriate treatments and recommendations in time. Such kind of a system will be a "Knight in the shining armor" for the farmers and can significantly increase the crop production on sustainable basis.

In order to overcome the problems of the manual, time consuming and costly approaches, several researchers have made efforts to automate this process of disease identification directly form images of leaves. These approaches are meant to detect various diseases in early stage so that appropriate treatment is done in time. Most of these techniques use machine learning and computer vision to classify images into different pre-defined classes. Machine Learning refers to the group of algorithms that learn from data and experience. Support Vector Machines (SVM), k-nearest neighbors (kNN), Artificial Neural Network (ANN) Decision Trees are some of the common learning algorithms extensively used for classification tasks. Mokhtar et al [4], used an SVM-based technique to detect diseases in tomatoes. SVM (Support Vector Machines) is a machine learning classification technique, which performs classification by maximizing the margin between classes. Dubey and Jalal used K-Means clustering technique for defect segmentation followed by classification using Multi-class SVM. The features used for classification are Global Color Histogram (GCH), Color Coherence Vector (CCV), Local Binary Pattern (LBP) and Completed Local Binary Pattern (CLBP) [5]. Dandawate and Kokare [6] 2015 and Raza et al [7] also used SVM based approaches for classification of plant diseases.

All the approaches discussed above are based on traditional Machine Learning techniques. These techniques are not fully automatic. In fact, one of the major drawbacks of Machine Learning is the need of manual intervention. Machine Learning techniques can work on small datasets but rely on manual hand crafted features designed by experts. Some of the features extensively used in computer vision are Local Binary Patterns (LBP), Complete Local Binary Pattern [8], Histogram of Gradients (HOG), HSV Histogram, Global Color Histogram and Gabor filters [9]. These handcrafted features are usually not robust, computationally intensive due to high dimensions, and are limited in number. Automated feature engineering on the other hand is more efficient and repeatable than manual feature engineering allowing us to build better predictive models faster.

Deep Learning is a subset of Machine Learning research, which has gained popularity in recent past. Deep Learning models are a special type of Artificial Neural Networks which learn (multiple levels of) representation by using a hierarchy of multiple layers. The biggest advantage of Deep Learning techniques is that they do not rely on hand-crafted features. Rather, these networks learn features while training without any human intervention. Recent advances in computer vision made possible by deep learning have paved the way for efficient and reliable visual systems that are extensively used in many areas like autonomous cars, medical image analysis, robotics etc. In recent years, Convolutional Neural Networks (CNNs), one of the successful deep learning algorithms, have dramatically achieved great success and won numerous contests in pattern recognition and Computer Vision [10]. They have shown excellent performance in many computer vision and machine learning tasks like image classification, object detection, speech recognition, natural language processing, medical image analysis etc. For example, in 2017, Brahimi et al, introduced deep learning as an approach for classifying tomato disease based on leaf images and achieved state-of-the-art results with a classification accuracy reaching up to 99%, easily outperforming the conventional methods [11].

The aim of this study is to utilize deep learning for classification of apple diseases. This study presents the following main contributions towards agriculture industry.

(1) There are mainly seven types of Kashmiri apples grown across different regions of the valley. Those seven types are Ambri, American Trel, Delicious, Maharaji, Hazratbali and Golden. The first part of this study is to build a large image dataset of healthy and diseased leaves covering almost all apple types and diseases. Having a sufficient sized dataset not only helps the current study but would potentially be helpful for future researches in the area.

(2) Another objective of this study is to provide accurate and reliable deep learning application which can identify and predict various types of apple disease by looking at the infected leaves. The proposed approach can significantly improve our existing disease management system.

The study should also help in introducing/promoting:

- Smart farming: Enabling the farmer to have firsthand knowledge about diseases in the earliest possible stage.
- Easy and early diagnosis of diseases.
- Enhancing the value of fruit disease detection.
- Enhance the apple economy through disease forecasting.

## 2. Major Apple Fruit Diseases of the Valley

Diseases and pests affecting the crops are one of the major shortcomings of horticulture in Kashmir. The most common and widely reported diseases and pests in Kashmir Valley are Apple Scab, Sooty blotch, Brown rot, Alternaria, Powdery mildew, canker, Red Mite, Sanjose scale, Wooly Aphid etc.

- *Apple scab* is a bacterial disease that affects both leaves and fruits. The symptoms of Apple scab appear as circular gray or brown corky spot on the surface. Severely affected leaves may turn yellow and fall.
- *Alternaria* is another bacterial disease that appear on leaves. The disease appears as round brown spot on the surface of the infected leaves.
- *Sooty blotch* is a fungal disease that appear as green, sooty or cloudy blotches on the surface of the fruit.
- *Flyspeck* is another fungal disease that appear as shiny round dots resembling fly excreta on the surface of the fruit. These dots are slightly raised from the surface and are present in groups.
- *Powdery mildew* disease appears on leaves and occurs as white or grey powder on the under surfaces of leaves. Fruits affected with powdery mildew fungus remain small and deformed.
- *Apple Mosaic* (ApMV)*:* Apple mosaic is a common viral disease found in India. Apple Mosaic appear as bright cream spots on spring leaves in spring season and can turn necrotic during summer.
- *Marssonina leaf blotch (MLB): MLB is a fungal disease which* appears in form of dark green circular patches on upper surface of leaf turning brown in due course.

3. **Work Methodology**

The working methodology of the proposed approach contains three components illustrated in Figure 1.

1. *Data collection and preparation:* Deep learning is all about data which serves as fuel in these learning models. Unfortunately, there is no appropriate sized dataset available that can be utilized for this study. Therefore, it is necessary to build a new dataset for the present study. Dataset preparation includes collection of infected/diseased leaf and fruit images of various apple varieties and categorizing them according to the type of disease.

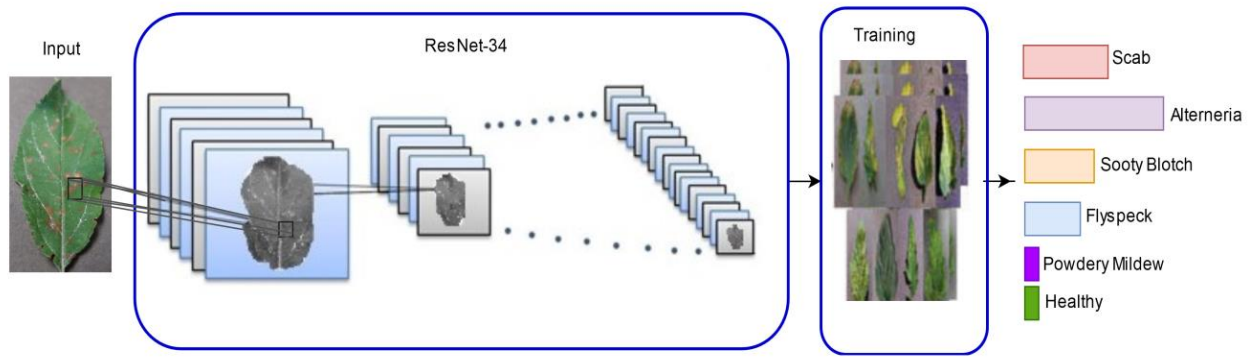

Figure 1: Overview of the proposed methodology

2. *Model development and training:* In this phase, we present a deep learning model which is a powerful Convolutional Neural Network (CNN) architecture for apple disease classification. The best thing about deep models is that they learn from data without any manual intervention. The features learned during training are much more useful and accurate than manual hand-crafted features. We use transfer learning to initialize the model parameters and then fine-tune on our prepared dataset. The complete details of the model and training is discussed in coming sections.
3. *Classification:* After training, the deep model is deployed on user machines. The users can upload an image of an infected leaf and the application will detect the type of disease that has infected the plant.

4. **Dataset Preparation**

There are around seven varieties of apples and eight types of diseases and pests commonly found in the valley. For this study, we have considered five commonly found diseases viz. *Scab, Alternaria, Apple Mosaic, Marssonina leaf blotch (MLB) and powdery mildew* only. Most of the data was collected from the apple orchards of Sher-e-Kashmir University of Agriculture Sciences and Technology, Kashmir (SKUAST-K) where they grow different varieties of fruits for educational and research purpose. Other than SKUAST-K, data collection was also carried out in various commercial apple growing areas of the valley like Baramulla, Shopian, Pulwama, Budgam etc.

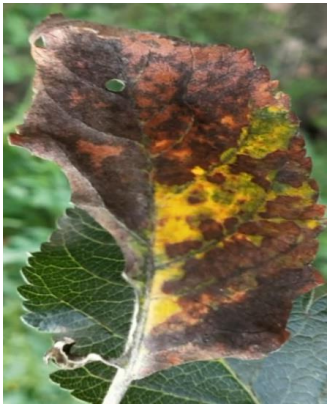 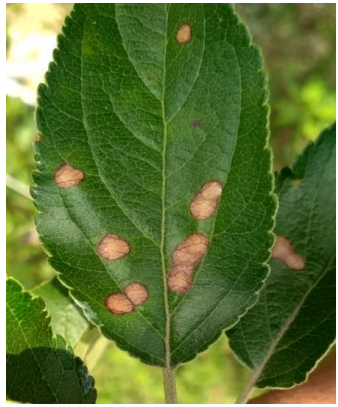 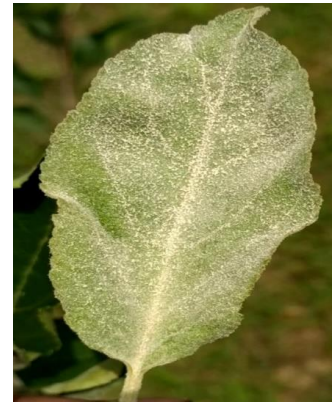

Scab                Alternaria              Powdery Mildew

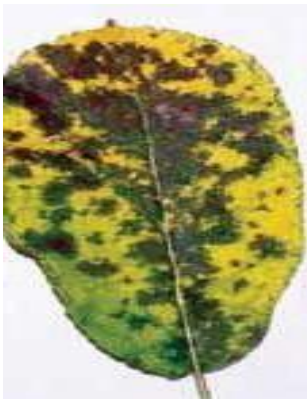 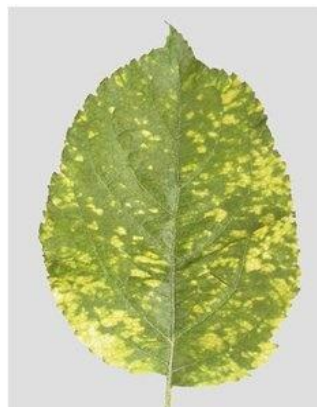 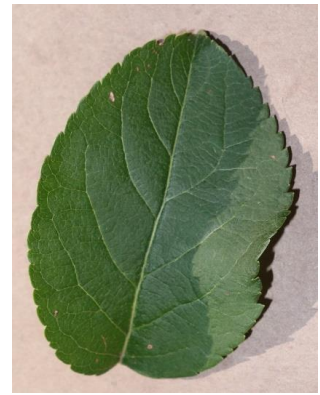

Marssonina Leaf Blotch     Apple Mosaic            Healthy

Figure 2: Sample images from the prepared Dataset

We collected around 8400 images of infected and healthy leaves. Most of the data collection was done during the months of June, July and August when maximum number of diseases are present on plants. All the images were captured manually using digital camera and mobile phones of different brands. The reason for using different capturing devices is to make sure that our dataset consists of images with variable illumination and quality so that our model generalizes well to the future unseen data. We expect our model to be tested and used in real environment where people use their mobile phones to capture images which are often blurry and poorly lit. Therefore, it is very important that our training set and testing set come from same distribution. Figure 2 shows some of the sample images from our prepared dataset. After capturing, all the images were resized to the dimension of 224 x 224 pixels with a resolution of 72 dpi. The images were then manually labelled with the help of two domain experts from SKUAST-K. The images were categorized into six classes according to their diseases. Table I below shows the summary of the prepared dataset.

*Table I: Dataset Summary*

| Disease | No. of Images |
| --- | --- |
| Scab | 1556 |
| Alternaria | 1550 |
| Apple Mosaic | 1300 |
| Marssonina Leaf Blotch (MLB) | 1312 |
| Powdery Mildew | 1356 |
| Healthy | 1350 |

The prepared dataset was then split into training set and validation set comprising of 70% and 30% of total data respectively.

## 5. Model Development and Training

In recent years, deep neural network techniques have shown tremendous performance in computer vision and pattern recognition tasks. One of the deep learning technique called as Convolutional Neural Network (CNN) is a neural network architecture with multiple hidden layers, which uses local connections known as local receptive field and weight-sharing for better performance and efficiency. The deep architecture helps these networks learn many different and complex features which a simple neural network cannot learn. CNN-based techniques have evolved as powerful visual models and achieved state-of-the-art performance in solving different problems of computer vision and pattern recognition like object detection and image classification [12]. In past few years, we have seen numerous CNN architectures starting from as few as 9 layers and going up to hundreds of layers. In this study, we use a modified pre-trained CNN model called ResNet-34. ResNet-34 is a 34-layer deep Convolutional Neural Network architecture developed by Microsoft Research Asia [13]. The reason for using pre-trained model is that it effectively reduces training time and also improves accuracy of models designed for tasks with minimum or inadequate training data [14]. The process of using a pre-trained model for a different task is also referred as Transfer Learning. The existing pre-trained model has to be modified according to the requirements before it can be used for a new task. For example, ResNet-34 has 1000 classes as the model is trained on ImageNet dataset which contains 1000 classes. So we replace the last layer of the ResNet-34 model with six classes. After model preparation, the modified pre-trained model is trained on target dataset. The architecture details of the modified ResNet-34 are as under.

ResNet is a deep CNN architecture which won 1st place in the ILSVRC 2015 classification competition with top-5 error rate of 3.57%. ResNet outperformed all its predecessor models both in accuracy as well as efficiency. ResNet has many variants like ResNet-34, ResNet-50, ResNet-101 etc. In this study, we use ResNet-34 variant which has 16 residual blocks and each block is 2 layers deep. Figure 3 shows the architecture of ResNet-34. The first layer of ResNet-34 is a Convolution layer which convolves an input image of size 224 x 224 with 64 kernels of size 7x7, producing 64 feature maps of size 112 x 112. The convolution operation is followed by batch-normalization and max-pooling operation. The first layer is followed by ResNet layers. In ResNet-34, there are four ResNet layers consists of 3, 4, 6 and 3 Residual Blocks respectively. A residual block consists of two convolution layers and a bypass connection as shown in Figure 3. Finally, the output of last ResNet layer is fed to a pooling layer followed by a 1000-way Fully Connected (FC) layer. For this study, we replaced the last 1000-way FC layer with 6-way FC layer because we have only 6 classes which represent the probabilities of each disease. In our case, we have six classes (Scab, Alternaria, *Apple Mosaic, Marssonina Leaf Blotch, Powdery mildew and Healthy*).

**Implementation and Training**

The proposed technique was implemented in Keras on top of Tensorflow 2.0 on a workstation with 16GB RAM, Intel Core i-5 9600k processor with RTX 2060 Super (8GB) graphics card. The model was trained on prepared dataset using stochastic gradient descent (SGD) with learning rate of 0.001, batch size of 8 and epoch value of 100. Figure 4 shows graph for training and validation losses of our proposed model which converged after around 50 epochs and the final validation accuracy reached up to 97%.

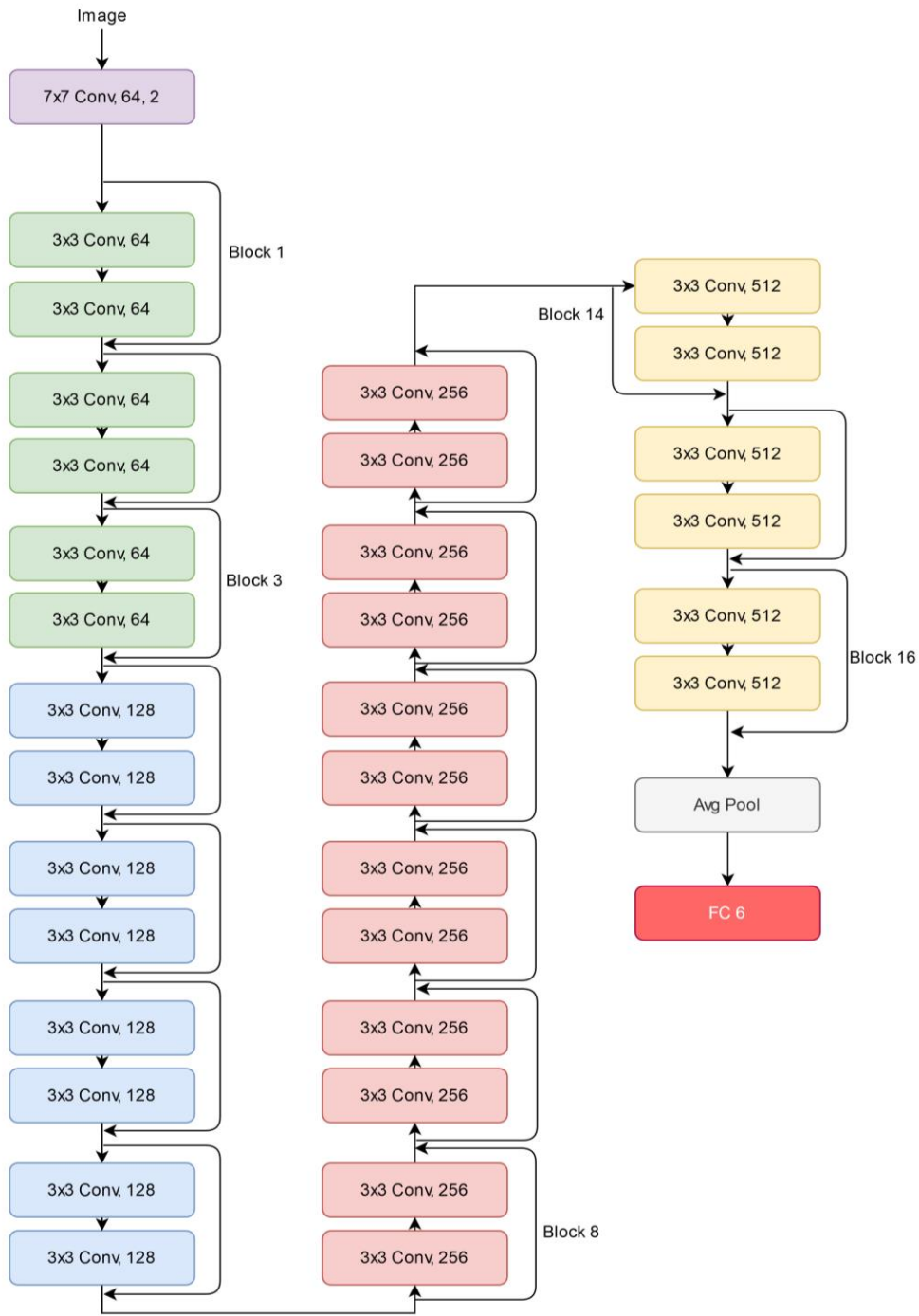

Figure 3: Architecture of the modified ResNet-34 model

## 6. Results and Discussion

The classification result of the proposed model on our prepared dataset was recorded and presented in the form of Confusion matrix (CM) in Table II and the graph is shown in Figure 5. Overall Accuracy, precision, recall, specificity and F-measure computed for each disease by formulae given below are summarized in Table III

$$\text{Accuracy} = \frac{No.\,of\ images\ correctly\ classified}{Total\ no.\,of\ images}$$

$$\text{Precision} = \frac{True\ Positives\ (TP)}{True\ Positives\ (TP) + False\ Positives\ (FP)}$$

$$\text{Recall} = \frac{True\ Positives\ (TP)}{True\ Positives\ (TP) + False\ Negatives\ (FN)}$$

$$\text{Specificity} = \frac{True\ Negatives\ (TN)}{True\ Negatives\ (TP) + False\ Positives\ (FP)}$$

$$\text{F-measure} = \frac{2 * Precision * Recall}{Precision + Recall}$$

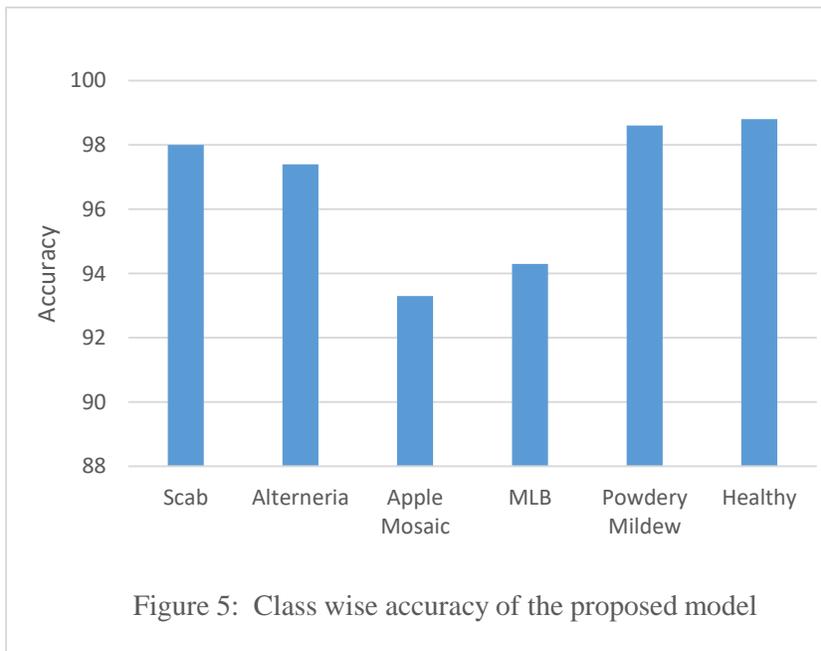

Figure 5: Class wise accuracy of the proposed model

These performance metrics are the top metrics used to measure the performance of classification algorithms. Our proposed model achieved an average accuracy of 97.2%, while as average precision, recall, specificity and F-measure (F1-Score) are 96.9%, 96.85%, 99.3% and 96.85 % respectively. The class-wise performance is presented in Table III. Healthy (normal) leaves are very easily distinguishable from diseased ones and a very good classification result is achieved for healthy class. Out of the five diseases, the performance for Powdery Mildew is slightly better with precision of 99.1%, recall 99.2%, Specificity 99.8% and F-measure 98.9%. For Apple Mosaic and Marssonina Leaf Blotch (MLB), our model recorded lowest accuracy of 93.3% and 94.3% respectively when compared to other classes. The low accuracy is mainly due to the reason that symptoms produced by different diseases may be very similar, and they may be present simultaneously. For example, the symptoms of MLB and scab may look similar at some stage which results in misclassification. However, one positive observation from the results is the precision (PPV) and recall (Sensitivity) values for top two diseases (Scab and Alterneria) of apple. Higher recall value means low false negative (FN) cases and low number of FN is an encouraging result.

**Table II: Classification Result of proposed Deep Model on prepared dataset**

| Disease | Predicted Class | | | | | | Accuracy |
| --- | --- | --- | --- | --- | --- | --- | --- |
| | Scab | Alternaria | Apple Mosaic | MLB | Powdery Mildew | Healthy | **97.2%** |
| Scab | **353** | 2 | 0 | 5 | 0 | 0 | 98% |
| Alternaria | 3 | **341** | 1 | 5 | 0 | 0 | 97.4% |
| Apple Mosaic | 4 | 2 | **280** | 3 | 9 | 2 | 93.3% |
| MLB | 8 | 7 | 2 | **283** | 0 | 0 | 94.3% |
| Powdery Mildew | 0 | 0 | 1 | 0 | **298** | 1 | 99.3% |
| Healthy | 0 | 2 | 2 | 0 | 0 | **346** | 98.8% |

The promising and encouraging results of deep learning approach in detection of diseases from leaf images indicate that deep learning has a greater role to play in disease detection and management in near future. Some limitation of this study can be overcome with more in depth analysis which is possible once more data becomes available.

*Table III: Class-wise precision, recall, Specificity and F-measure of proposed model*

| Class | Precision (%) | Recall (%) | Specificity (%) | F-measure (%) |
|---:|:---:|:---:|:---:|:---:|
| *Scab* | 95.9 | 98 | 99 | 96.9 |
| *Alternaria* | 96.3 | 97.4 | 99.1 | 96.8 |
| *Apple Mosaic* | 97.9 | 93.3 | 99.6 | 95.5 |
| *MLB* | 95.6 | 94.3 | 99.2 | 94.9 |
| *Powdery Mildew* | 97 | **99.3** | 99.4 | 98.1 |
| *Healthy* | **99.1** | 98.8 | **99.8** | **98.9** |
| *Average* | **96.9** | **96.85** | **99.3** | **96.85** |

**Conclusion**

In this study we first prepared a dataset of healthy and infected apple leaves which we collected from various orchards located of the Kashmir valley. We then trained a deep learning model initialized using transfer learning for automatic apple disease identification and classification on our prepared dataset. The results obtained by the proposed approach were promising, reaching an accuracy of around 97%. In future, our aim is to develop a complete disease detection and recommendation system that will assist farmers to take timely actions upon detection of a disease. The localization of infected region in an infected area will help users by giving them information about the disease without the intervention of agriculture experts. Moreover, the system can send the information along with the location details to the data server where this information can be used for disease analysis, fruit production analysis and disease forecasting. Moreover, location details can help the experts analyze the spread of a particular disease area wise so that the farmers can be cautioned in advance about the spread of any disease and hence avoid any catastrophic damage.